\ifdefined\pdftexversion\pdfoutput=1\fi

\newcommand{\authorblock}{%
Woosik Kim$^{1}$ \quad Wonhyeok Choi$^{2}$ \quad Sunghoon Im$^{1,\dagger}$\\[3pt]
{\normalsize $^{1}$KAIST, Daejeon, Republic of Korea \qquad $^{2}$DGIST, Daegu, Republic of Korea}\\[2pt]
{\normalsize\tt woosik@kaist.ac.kr \quad smu06117@dgist.ac.kr \quad im@kaist.ac.kr}%
\thanks{$^{\dagger}$Corresponding author.}%
}

\documentclass[letterpaper, 10pt, conference]{ieeeconf}
\IEEEoverridecommandlockouts
\usepackage[utf8]{inputenc}
\usepackage[T1]{fontenc}
\usepackage{microtype}
\usepackage{amsmath,amssymb,amsfonts}
\usepackage{graphicx}
\usepackage{booktabs}
\usepackage{xcolor}
\usepackage{xspace}
\usepackage{float}
\usepackage{multirow}
\usepackage{rotating}
\usepackage[hidelinks]{hyperref}

\newcommand{\subsetname}{\textsc{Expressive-51}\xspace}
\newcommand{\method}{\textsc{Cantabile}\xspace}

\definecolor{tabgain}{HTML}{1F5FA8}
\definecolor{tabloss}{HTML}{B03A2E}
\newcommand{\gain}[1]{{\scriptsize\color{tabgain}#1}}
\newcommand{\loss}[1]{{\scriptsize\color{tabloss}#1}}

\title{\LARGE \bf CANTABILE: Learning Expressive Dynamics for Robotic Piano Performance}

\providecommand{\authorblock}{Anonymous Author(s)}
\author{\authorblock}

\begin{document}
\maketitle
\thispagestyle{empty}
\pagestyle{empty}

\begin{abstract}
Robotic piano playing has emerged as a standard benchmark for dexterous bimanual manipulation, yet progress on it has been measured almost entirely by note accuracy---which keys are pressed (pitch) and when (onset)---leaving the musical dynamics essential for expressive performance neither rewarded nor evaluated.
We propose \method, a dynamics-aware framework for robotic piano performance that (i)~closes the score-to-contact loop by conditioning the policy on upcoming velocity goals and mapping each key's angular velocity at onset back to MIDI velocity, (ii)~couples a velocity-fidelity reward with an onset-coverage reward, so that dynamics cannot be improved by omitting difficult notes, and (iii)~refines a frozen dynamics-aware base policy with an $\alpha$-scaled, finger-only residual that localizes strike-intensity adaptation away from nominal note execution.
On \subsetname, a dynamics-rich 51-song subset of RoboPianist, \method raises Velocity F1---jointly measuring pitch, onset, and intensity within a $\pm8$ MIDI-velocity tolerance---from 0.06 to 0.34 over the RoboPianist baseline, improves all 51 songs, more than halves matched-note velocity error, and reduces log-mel distance to reference audio by 8\%.
Intensity-randomized training further enables runtime control of performance intensity without retraining.
\end{abstract}

\section{Introduction}
\label{sec:intro}

Dexterous manipulation is a long-standing challenge in robot learning~\cite{andrychowicz2020learning,chen2021system}, and robotic piano playing has emerged as one of its most demanding testbeds: high-dimensional, contact-rich, and bimanual~\cite{zakka2023robopianist}. Successful performance requires coordinating two multi-fingered hands across 88 keys, planning long sequences of movements, and executing precisely timed contacts.
Recent work has substantially advanced this problem through reinforcement learning in simulation~\cite{zakka2023robopianist}, multi-song generalization~\cite{zhao2025rp1m,qian2025pianomime,chen2025omnipianist}, sim-to-real transfer~\cite{zeulner2025realpiano,xie2026handelbot}, and motion-level realism~\cite{wang2024furelise}. These advances have broadened the repertoire, deployability, and motion realism of robotic piano playing. Yet the acoustic character of the resulting performance remains weakly represented in standard benchmarks.

\begin{figure}[t]
\centering
\includegraphics[width=\linewidth]{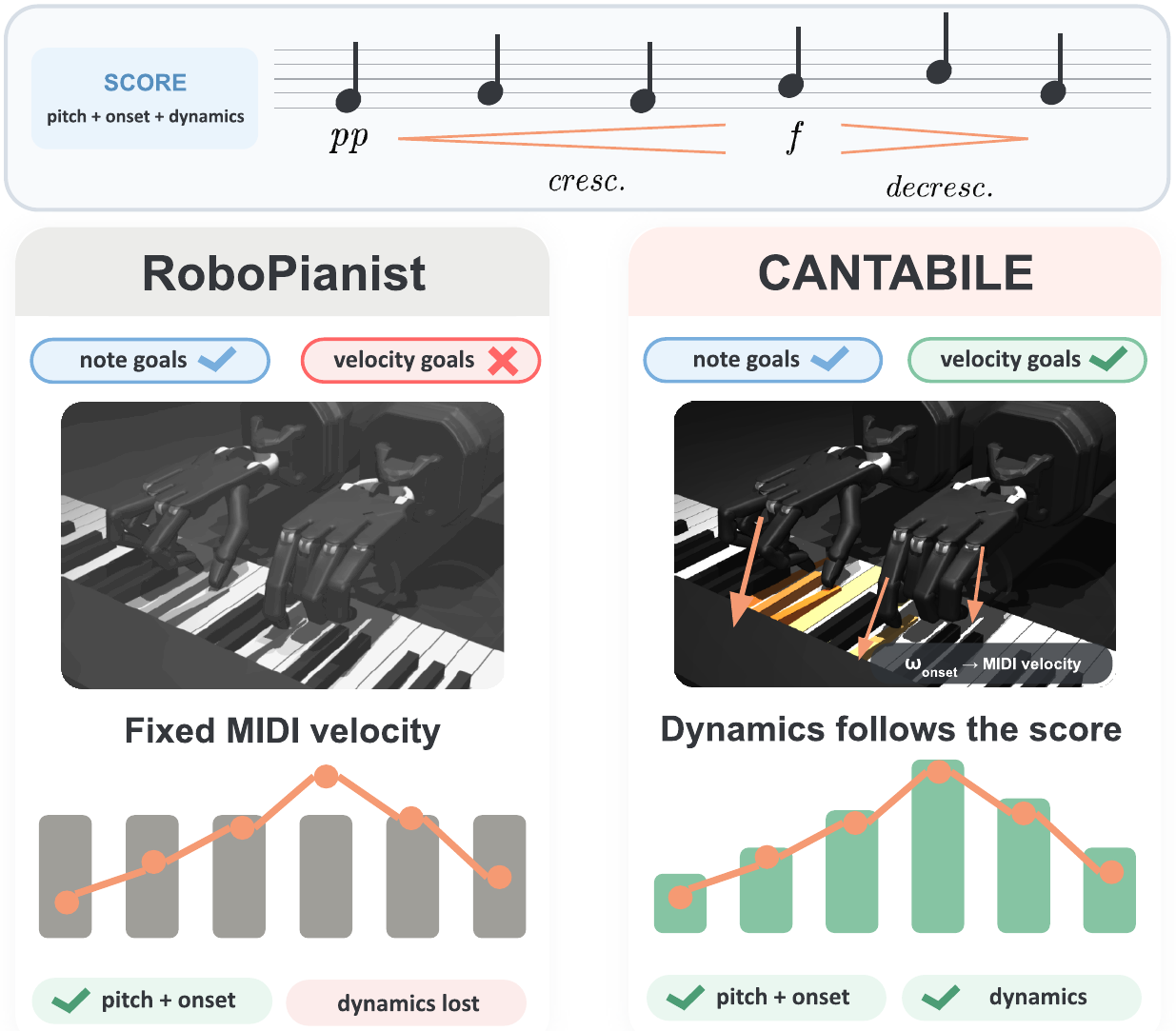}
\caption{\textbf{Note accuracy alone cannot see musical dynamics.} A score specifies pitch, onset \emph{and} intensity, but the note-accuracy objective of RoboPianist~\cite{zakka2023robopianist} (\emph{left}) scores only pitch and onset, rendering every keystroke at a fixed MIDI velocity. \method (\emph{right}) senses each key's speed at the instant it is struck, maps it to MIDI velocity, and rewards loudness that follows the notated dynamics.}
\label{fig:teaser}
\end{figure}

An essential component of musical performance is dynamics: the pattern of intensity across individual notes.
Piano loudness is strongly influenced by hammer impact speed~\cite{russell1998hammer}, which is induced by the key's velocity during a strike.
Thus, performances with identical notes and timing can nonetheless yield different acoustic impressions.
Specifically, the robotic piano-playing task~\cite{zakka2023robopianist} evaluates performance primarily through key-press accuracy and renders every detected keystroke at a fixed MIDI velocity (Figure~\ref{fig:teaser}).
Consequently, differences in key-impact speed are neither reflected in the rendered output nor captured by the task objective.

\begin{figure*}[!t]
\centering
\includegraphics[width=\textwidth]{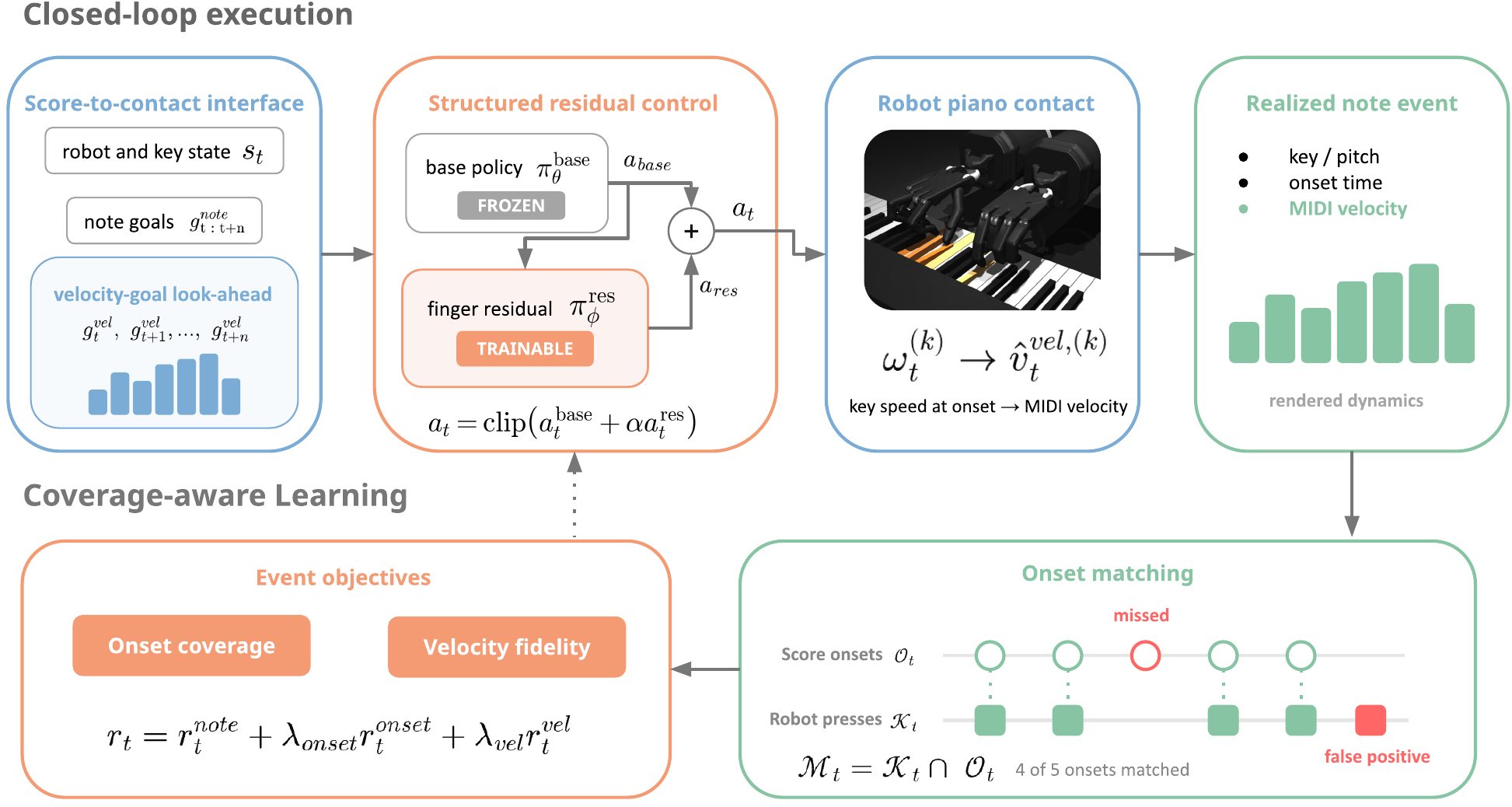}
\caption{\textbf{\method overview.} \emph{Top, closed-loop execution:} the score-to-contact interface feeds the robot and key state, plus the upcoming velocity goals, to a frozen base policy and a trainable $\alpha$-scaled finger residual, and reads the key's angular speed at onset back as a MIDI velocity (Eq.~\eqref{eq:vel-map}). \emph{Bottom, coverage-aware learning:} executed key presses are matched against the score's onsets within the same control step, and the matched set $\mathcal{M}_t$ drives the onset-coverage and velocity-fidelity objectives, which update the finger residual only.}
\label{fig:overview}
\end{figure*}

We argue that addressing this limitation requires a dynamics-aware framework that treats velocity as a first-class control objective throughout the learning and evaluation pipeline.
Expressive-performance rendering models~\cite{borovik2023scoreperformer,jeong2019virtuosonet,oore2020feeling} explicitly model MIDI velocity, but treat it as a symbolic performance parameter rather than a consequence of robot-piano contact.
In an embodied setting, dynamics is an event-level outcome fixed at onset (\textit{i.e.}, the moment a key is pressed), so the formulation must anticipate the target, measure the realized intensity, and account for missed or mistimed notes.
Otherwise, learning from and evaluating only successfully played events can make a selective, low-coverage performance appear dynamically accurate.
Dynamics fidelity and event coverage are therefore inseparable in both learning and evaluation.

To operationalize this principle, we propose \method, a framework that closes the loop between score-level dynamics and embodied execution.
\method represents velocity consistently from score-specified targets to the key motion realized at each note onset, and couples velocity fidelity with onset coverage so that gains are not achieved simply by playing fewer notes.
It further separates fine-grained dynamics adaptation from nominal note execution through localized finger corrections to a frozen piano-playing policy.
Together, these components make dynamics measurable, learnable, and controllable while limiting interference with the base policy's note-playing behavior.
For repertoire-scale evaluation, we curate \subsetname, a 51-piece suite selected from the RoboPianist~\cite{zakka2023robopianist} repertoire for substantial variation in notated MIDI velocity, and introduce Velocity F1, which jointly assesses pitch (\textit{i.e.}, note accuracy), onset, and intensity.

Our contributions are summarized as follows:
\begin{itemize}\setlength\itemsep{1pt}
  \item We introduce a \emph{score-to-contact dynamics interface} that conditions control on upcoming velocity and maps key-onset motion to MIDI velocity and audio.
  \item We propose \emph{coverage-aware event learning}, coupling velocity supervision with onset coverage to prevent selective note omission.
  \item We develop a \emph{coarse-to-fine controller} that refines a frozen dynamics-aware base with a finger-only residual while preserving nominal coordination.
  \item We present \subsetname and \emph{Velocity F1} for joint pitch--onset--intensity evaluation, and enable runtime intensity control via intensity randomization.
\end{itemize}

\section{Related Work}
\label{sec:related}

\paragraph{Robotic piano playing}
Learning-based piano playing predates the current benchmark~\cite{xu2022touch}, where a 24-key setup rewarded the difference between played and notated key velocity; the field then consolidated around RoboPianist~\cite{zakka2023robopianist}, which trains bimanual Shadow Hands per piece and scores them by key-press F1.
Follow-up work has scaled the repertoire---RP1M~\cite{zhao2025rp1m} through a million machine-generated expert trajectories, PianoMime~\cite{qian2025pianomime} by distilling YouTube performances, OmniPianist~\cite{chen2025omnipianist} with a flow-matching generalist---transferred policies onto real hardware~\cite{zeulner2025realpiano,xie2026handelbot}, made the motion human-like~\cite{wang2024furelise}, and added real-time accompaniment of a human pianist~\cite{wang2024cooperative}.
Despite these different goals, the benchmark and the work built on it measure success by the same criterion---which keys were pressed, and when---and MIDI velocity appears in neither their reward nor their evaluation metric.
Concurrent work~\cite{liang2026expressive} adds a score-velocity matching reward to per-piece policies for a single dexterous hand and deploys them on a digital piano.
That work demonstrates human-like fingering and real-piano deployment, whereas we target the bimanual benchmark at repertoire scale and score dynamics jointly with onset accuracy.

\begin{figure*}[!t]
\centering
\includegraphics[width=\textwidth]{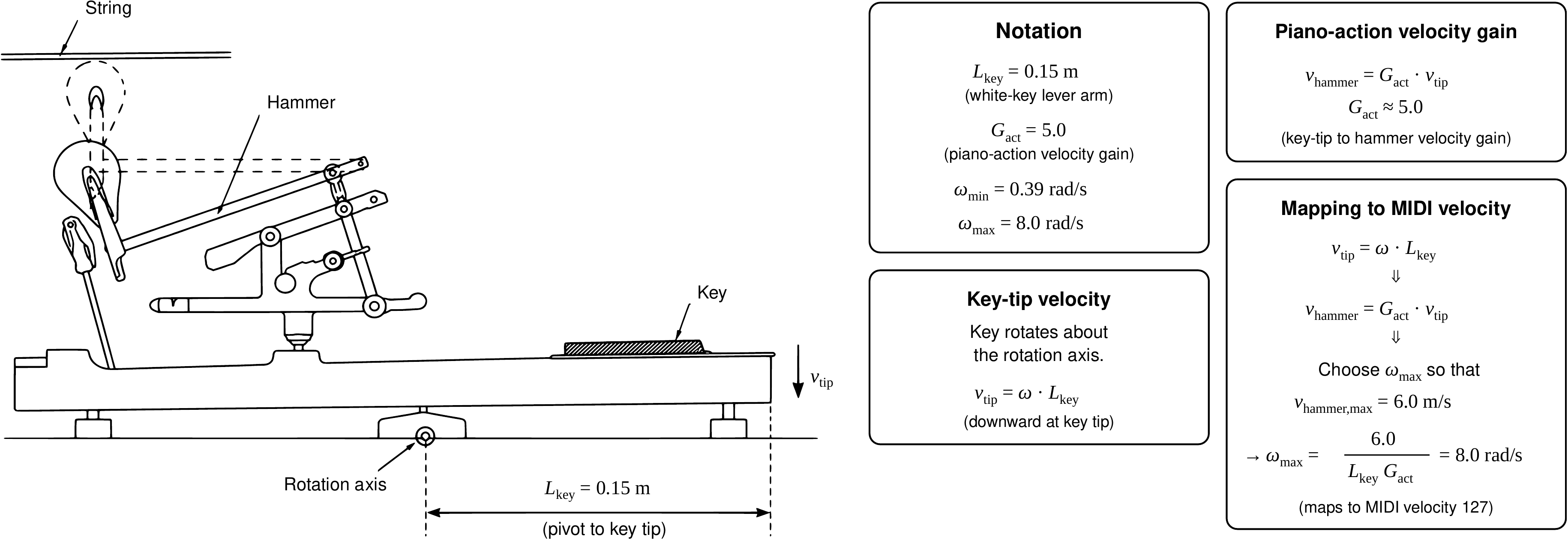}
\caption{\textbf{Physical basis of velocity sensing.}
Key angular velocity $\omega$ produces key-tip speed $v_{\mathrm{tip}} = \omega\,L_{\mathrm{key}}$, which the piano action amplifies to hammer velocity by the ratio $G_{\mathrm{act}}$, and which Eq.~\eqref{eq:vel-map} then maps to MIDI velocity.}
\label{fig:vel-sensing}
\end{figure*}

\paragraph{Expressive performance rendering}
In music information retrieval, by contrast, velocity has long been treated as central to expressive performance~\cite{cancino2018computational}: rendering a score expressively means choosing each note's loudness, and velocity has accordingly been modeled as a first-class output---autoregressively by Performance RNN~\cite{oore2020feeling}, hierarchically by VirtuosoNet~\cite{jeong2019virtuosonet}, and with diffusion by DExter~\cite{zhang2024dexter}. ScorePerformer~\cite{borovik2023scoreperformer} established the evaluation conventions we adopt directly: per-note error metrics computed on matched note events.
None of these models controls a physical system---the velocities they predict are symbolic outputs, not the result of motor commands---and we bring their evaluation methodology into the embodied setting, where velocity must be realized through contact.

\paragraph{Residual reinforcement learning}
Residual RL lets a learned corrective term improve a controller that is hard to train from scratch, from fixed initial controllers~\cite{silver2018residual} to hand-engineered controllers for contact-rich manipulation~\cite{johannink2019residualrl,davchev2022residual}; later formulations recast it as policy customization---preserving a prior policy while satisfying an additional downstream requirement~\cite{li2023residualq,wang2025residualpg}.
In the piano domain, the pattern reappears in PianoMime~\cite{qian2025pianomime}, correcting imitation errors of per-song experts, and in HandelBot~\cite{xie2026handelbot}, closing the sim-to-real gap. In all of these, the residual serves the \emph{same objective} its base already pursues; ours carries a \emph{new} one---expressive dynamics---while the frozen base retains note accuracy, and the structural constraints ($\alpha$-scaled, finger-only) keep the new objective from overwriting the old.

\section{Method}
\label{sec:method}

\label{sec:overview}
\method connects score-specified intensity to its physical realization through three components (Figure~\ref{fig:overview}): a score-to-contact interface that provides anticipatory targets and measures key-onset motion (Section~\ref{sec:env}); a coverage-aware objective combining velocity supervision with onset coverage (Section~\ref{sec:mdp}); and a finger-only residual controller that preserves nominal note execution (Section~\ref{sec:residual}).

\subsection{Score-to-Contact Dynamics Interface}
\label{sec:env}

Each score onset on key $k$ at control step $t$ carries a target MIDI velocity $g_t^{\mathrm{vel},(k)}\in\{1,\ldots,127\}$, which encodes the intended strike intensity. \method links this score-level target to physical execution through a score-to-contact dynamics interface.

\paragraph{Velocity-goal look-ahead}
Let $o_t^{\mathrm{rp}}$ denote the standard RoboPianist observation, comprising the joint robot--piano state $s_t$ and note-goal look-ahead $g^{\mathrm{note}}_{t:t+n_{\mathrm{note}}}$.
For each future step $t+i$, we construct $g_{t+i}^{\mathrm{vel}}\in[0,1]^{88}$, whose $k$-th entry is $g_{t+i}^{\mathrm{vel},(k)}/127$ for a key with a target onset at that step and zero otherwise. The dynamics-aware observation is
\begin{equation}
o_t=\left[o_t^{\mathrm{rp}},\,g_t^{\mathrm{vel}},\ldots,g_{t+n_{\mathrm{vel}}}^{\mathrm{vel}}\right].
\label{eq:velocity-observation}
\end{equation}
The base task's horizon is $n_{\mathrm{note}}{=}10$; we use $n_{\mathrm{vel}}{=}2$, selected in Section~\ref{sec:robustness}.
Because strike intensity is determined at contact, this look-ahead lets the policy adjust its motion before the onset. It is the only feature added to the standard policy observation. Control remains closed-loop through the robot and key states in $o_t^{\mathrm{rp}}$; the realized MIDI velocity is an event-level output used for rendering and, during training, reward computation.

\paragraph{Onset and velocity sensing}
We detect an executed onset when key $k$ first crosses the pressed-state threshold at step $t$ and read its angular speed $\omega_t^{(k)}$ from the simulator.
We then map this speed to a realized MIDI velocity $\hat{v}_t^{\mathrm{vel},(k)}$ (Figure~\ref{fig:vel-sensing}):
\begin{equation}
\hat{v}_t^{\mathrm{vel},(k)} \;=\; \mathrm{clip}\!\left(
\frac{|\omega_t^{(k)}|-\omega_{\min}}
{\omega_{\max}-\omega_{\min}}
\cdot126,0,126\right)+1.
\label{eq:vel-map}
\end{equation}
The mapping is fixed across all policies. We set $\omega_{\min}=0.39$~rad/s as the empirically determined lower endpoint for a registered onset in the simulator. For the upper endpoint, key-tip and hammer velocities satisfy $v_{\mathrm{tip}}=|\omega|L_{\mathrm{key}}$ and $v_{\mathrm{hammer}}=G_{\mathrm{act}}v_{\mathrm{tip}}$. Using $L_{\mathrm{key}}=0.15$~m, the approximate hammer-to-key-tip speed ratio $G_{\mathrm{act}}=5$~\cite{askenfelt1991motion}, and a maximum hammer velocity of $6$~m/s~\cite{russell1998hammer} gives $\omega_{\max}=6.0/(0.15\times5.0)=8.0$~rad/s. We linearly interpolate between these endpoints as a monotonic approximation, consistent with the reported relationship between hammer velocity and tone amplitude~\cite{palmer1991amplitude}.


\subsection{Coverage-Aware Event Learning}
\label{sec:mdp}

Velocity supervision is sparse because it exists only when the robot produces an onset. Optimizing velocity error on successful notes alone is therefore exploitable: the policy can avoid a negative velocity reward by omitting a difficult onset. We address this failure mode by coupling matched-onset velocity fidelity with onset coverage.

Let $\mathcal{K}_t$ be the set of keys newly pressed by the robot at control step $t$, $\mathcal{O}_t$ the set of score onsets at that step, and $\mathcal{M}_t=\mathcal{K}_t\cap\mathcal{O}_t$ their matched set. Both score and executed onsets are discretized at the control rate; a match requires the same key in the same control step, with no additional temporal window. For a matched onset, we measure velocity agreement with the Gaussian tolerance $\rho(e)=\exp\!\left[\ln(0.05)(e/20)^2\right]$ of the velocity error $e$, so that $\rho(0)=1$ and $\rho(\pm20)=0.05$. Denoting the matched-pair value by $\rho_{t,k}=\rho\big(\hat{v}_t^{\mathrm{vel},(k)}-g_t^{\mathrm{vel},(k)}\big)$, the velocity-fidelity reward is
\begin{equation}
r_t^{\mathrm{vel}}=
\frac{1}{\max(|\mathcal{K}_t|,1)}
\sum_{k\in\mathcal{M}_t}
\left(2\rho_{t,k}-1\right).
\label{eq:vel-reward}
\end{equation}
A well-matched velocity receives a positive reward, a poorly matched one is penalized, and a step with no executed onset receives zero. Because a missed target also receives zero under Eq.~\eqref{eq:vel-reward}, this term alone can favor omission. We therefore introduce the onset-coverage reward
\begin{equation}
r_t^{\mathrm{onset}}=
\frac{|\mathcal{M}_t|}{\max(|\mathcal{O}_t|,1)},
\label{eq:onset-reward}
\end{equation}
which rewards the fraction of target onsets realized at the correct step. The complete training reward is
\begin{equation}
r_t=r_t^{\mathrm{note}}
+\lambda_{\mathrm{vel}}r_t^{\mathrm{vel}}
+\lambda_{\mathrm{onset}}r_t^{\mathrm{onset}}.
\label{eq:composite-reward}
\end{equation}
Here $r_t^{\mathrm{note}}$ denotes the unmodified RoboPianist task reward, which combines target-key actuation, false-press avoidance, sustain-pedal tracking, fingering guidance, and energy regularization (Table~\ref{tab:rewards}). We retain its original formulation and coefficients, and use $(\lambda_{\mathrm{vel}},\lambda_{\mathrm{onset}})=(0.2,0.5)$; the contribution and sensitivity of the two event terms are evaluated in Sections~\ref{sec:ablation} and~\ref{sec:robustness}.

\begin{table}[!t]
\centering
\caption{RoboPianist note-playing reward $r_t^{\mathrm{note}}$~\cite{zakka2023robopianist}. Here, $\rho(\cdot)$ is the Gaussian tolerance of Eq.~\eqref{eq:vel-reward} applied to each term's own error, with that term's original margin; $\mathcal{G}_t$ and $\mathcal{P}_t$ the target and pressed-key sets; $\Delta q_k$ and $\Delta q^{\mathrm{sus}}_t$ the key-depth and pedal errors; $p_f^{\mathrm{tip}}$ and $p_f^{\mathrm{key}}$ the fingertip and assigned-key positions; and $u_j,\dot q_j$ the actuator torque and joint velocity. A constant forearm-collision term is omitted.}
\label{tab:rewards}
\renewcommand{\arraystretch}{1.2}
\resizebox{\linewidth}{!}{%
\scriptsize
\begin{tabular}{ll}
\toprule
Term & Equation \\
\midrule
Key press &
$\frac{1}{2}\mathrm{mean}_{k\in\mathcal{G}_t}\rho(\Delta q_k)
+\frac{1}{2}\mathbf{1}
[\mathcal{P}_t\setminus\mathcal{G}_t=\varnothing]$  \\ 
Sustain & 
$\rho(\Delta q^{\mathrm{sus}}_t)$ \\
Fingering &
$\mathrm{mean}_f
\rho\!\left(
\lVert p_f^{\mathrm{tip}}-p_f^{\mathrm{key}}\rVert
\right)$ \\
Energy &
$-c_E\sum_j|u_j\dot q_j|$ \\
\bottomrule
\end{tabular}}
\end{table}

\paragraph{Runtime intensity commands}
As an optional controllability variant, we train the end-to-end dynamics policy with episode-level intensity randomization.
At each episode, we sample a global command $c\sim\mathcal{U}[0.5,1.5]$ and replace each nonzero target
velocity with $g_t^{\mathrm{cmd},(k)}=\mathrm{clip}(c\,g_t^{\mathrm{vel},(k)},1,127)$.
The commanded velocities are used in both the velocity-goal look-ahead and the velocity-fidelity reward.
At inference, varying $c$ changes the requested intensity without updating the policy parameters (Section~\ref{sec:controllability}).

\begin{table*}[!t]
\centering
\caption{\textbf{Main comparison on \subsetname} (mean over the 51 songs; one deterministic rollout per song and method; all methods compute-matched at $11{\times}10^6$ steps). Relative changes are against RoboPianist~\cite{zakka2023robopianist}, \textcolor{tabgain}{blue} where the metric improves and \textcolor{tabloss}{red} where it regresses.}
\label{tab:main}
\setlength{\tabcolsep}{11pt}
\begin{tabular}{lcccccc}
\toprule
Method & Note F1 $\uparrow$ & Onset F1 $\uparrow$ & Vel MAE $\downarrow$ & Vel F1 $\uparrow$ & Mel-L2 $\downarrow$ & Loud.\ corr $\uparrow$ \\
\midrule
RoboPianist~\cite{zakka2023robopianist} & \textbf{0.779} & 0.501 & 37.1 & 0.063 & 17.2 & 0.569 \\
End-to-End ($\pi^{\mathrm{vel}}$)  & 0.761~\loss{$-2.3\%$} & \textbf{0.536}~\gain{$+7.0\%$} & 19.8~\gain{$-47\%$} & 0.260~\gain{$+313\%$} & \textbf{15.5}~\gain{$-9.9\%$} & 0.628~\gain{$+10\%$} \\
Residual on RoboPianist            & 0.774~\loss{$-0.6\%$} & 0.534~\gain{$+6.6\%$} & 30.6~\gain{$-18\%$} & 0.198~\gain{$+214\%$} & 17.3~\loss{$+0.6\%$} & 0.581~\gain{$+2.1\%$} \\
\midrule
Residual on $\pi^{\mathrm{vel}}$   & 0.758~\loss{$-2.7\%$} & 0.533~\gain{$+6.4\%$} & \textbf{17.2}~\gain{$-54\%$} & \textbf{0.342}~\gain{$+443\%$} & 15.8~\gain{$-8.1\%$} & \textbf{0.649}~\gain{$+14\%$} \\
\bottomrule
\end{tabular}
\end{table*}

\subsection{Structure-Preserving Residual Control}
\label{sec:residual}

Jointly optimizing note execution and strike intensity over the full action space can perturb the coordination already acquired for key selection, hand transport, and onset timing. \method therefore adopts a two-stage, coarse-to-fine control scheme.

In the first stage, we train a dynamics-aware policy $\pi_\theta^{\mathrm{vel}}$ using the observation $o_t$ defined in Eq.~\eqref{eq:velocity-observation} and the composite reward in Eq.~\eqref{eq:composite-reward}.
In the second stage, we freeze this policy and use it as the base policy $\pi_\theta^{\mathrm{base}}$. A residual policy then receives both $o_t$ and the base action $a_t^{\mathrm{base}}$:
\begin{equation}
\begin{aligned}
a_t^{\mathrm{base}}
&\sim
\pi_\theta^{\mathrm{base}}
(\cdot\mid o_t),~~
a_t^{\mathrm{res}}
\sim
\pi_\phi^{\mathrm{res}}
(\cdot\mid o_t,a_t^{\mathrm{base}}),\\
a_t
&=
\mathrm{clip}\!\left(
a_t^{\mathrm{base}}
+\alpha\,
(m_f\odot a_t^{\mathrm{res}}),
a_{\min},a_{\max}
\right).
\end{aligned}
\label{eq:residual}
\end{equation}
Here, $m_f$ is a binary mask that retains only the finger-action dimensions, and $\alpha$ bounds the magnitude of the residual correction. Conditioning the residual on $a_t^{\mathrm{base}}$ allows it to refine the imminent action rather than infer a correction from $o_t$ alone.

The same construction applies with the RoboPianist policy $\pi^{\mathrm{rp}}$ as the frozen base, a variant we evaluate in Section~\ref{sec:capability}.
This structure separates the two control roles: the frozen base provides the nominal hand coordination and note execution, whereas the residual makes bounded, finger-localized corrections to strike intensity. The base policy remains fixed during the second stage, and only the residual policy is optimized using the composite reward in Eq.~\eqref{eq:composite-reward}. Both stages use Soft Actor-Critic (SAC) \cite{haarnoja2018sac} with automatic entropy-temperature tuning~\cite{haarnoja2018sacapps}.

\section{Experiments}
\label{sec:experiments}
Under the setup in Section~\ref{sec:setup}, we compare \method with compute-matched baselines in dynamics fidelity and note accuracy (Section~\ref{sec:capability}), ablate its event rewards and residual design (Section~\ref{sec:ablation}), evaluate generalization across runtime intensity commands (Section~\ref{sec:controllability}), and test robustness to design and evaluation choices (Section~\ref{sec:robustness}).

\subsection{Experimental Setup}
\label{sec:setup}

\paragraph{Evaluation songs}
We curate \subsetname, a 51-piece subset of the 150-piece repertoire of the robotic piano-playing task~\cite{zakka2023robopianist}, for meaningful evaluation of musical dynamics. In the full repertoire, 23 pieces use a constant MIDI velocity, while another 38 use no more than four distinct velocity levels, providing limited variation for distinguishing dynamics-aware methods. We therefore retain pieces whose ground-truth velocity sequence has a standard deviation of at least 10. We additionally exclude any piece containing notes with MIDI velocity $\leq 4$, which are effectively inaudible ($-53$ dBFS) when rendered with the benchmark soundfont.

\paragraph{Implementation details}
We train one policy per song using SAC~\cite{haarnoja2018sac} in the MuJoCo-based piano environment introduced in RoboPianist~\cite{zakka2023robopianist}.
The actor and critic use three-layer MLPs with 256 GELU units per layer; the learning rate, batch size, replay capacity, and discount factor are $3\times10^{-4}$, 256, $10^6$, and 0.8, respectively. The control rate is 20\,Hz.
RoboPianist~\cite{zakka2023robopianist}, which we retrain in this setup, and the end-to-end dynamics policy are trained for $11\times10^6$ environment steps. \method trains the dynamics-aware base for $8\times10^6$ steps and the finger-only residual for an additional $3\times10^6$ steps, matching the total interaction budget; the residual on RoboPianist follows the same schedule, with $\pi^{\mathrm{rp}}$ as the $8\times10^6$-step base. All methods receive the same observation, including the velocity-goal look-ahead; RoboPianist's policy $\pi^{\mathrm{rp}}$ differs only in its reward, omitting the two event-level objectives.
At evaluation, we use the mean policy action and a fixed initial state. Under this deterministic evaluation setting, repeated rollouts of the same checkpoint are identical; we therefore report one rollout per song and method.

\paragraph{Evaluation metrics}
Note accuracy is measured with the standard key-press F1, which we write as note F1.
For dynamics, \emph{velocity MAE} is the mean absolute difference between performed and target MIDI velocity over the matched onsets $\mathcal{M}$ of Section~\ref{sec:mdp}, accumulated over the episode.
Because it is computed only over the notes struck in time, a policy that matches few onsets can still score well on it.

Our primary metric therefore evaluates every note jointly on timing and loudness.
\emph{Velocity F1} counts a keystroke as a true positive only if it lands in the same control step as a score onset \emph{and} within $\tau$ MIDI-velocity units of its velocity goal $g_t^{\mathrm{vel},(k)}$:
\begin{equation}
  P_{\mathrm{vel}} = \frac{\mathrm{TP}_\tau}{|\mathcal{K}|}, \qquad
  R_{\mathrm{vel}} = \frac{\mathrm{TP}_\tau}{|\mathcal{O}|},
\label{eq:vel-f1}
\end{equation}
with precision over \emph{all} robot keystrokes and recall over \emph{all} score onsets, so spurious and missed notes are penalized rather than ignored.
We set $\tau{=}8$, half a notated dynamic level: \emph{ppp} to \emph{fff} divide the MIDI range into eight levels of roughly 16 units, so an accepted note stays at its marked dynamic. The tolerance is stricter than the 10.8-unit mean velocity MAE we measure between 299 pairs of professional performances of the same piece in MAESTRO~\cite{hawthorne2019maestro}. Even reproducing pianos realize a specified velocity only approximately~\cite{goebl2003measurement}.
Section~\ref{sec:robustness} verifies that the ranking of methods does not depend on this choice.
Removing the loudness condition ($\tau\to\infty$) recovers a purely timing-level \emph{Onset F1}, reported alongside to separate \emph{when} from \emph{how-loud} errors---mirroring the note / note-with-velocity convention of piano transcription evaluation~\cite{raffel2014mireval,hawthorne2018onsets,kong2021high}.

Finally, to measure what a listener would hear, we render the reference score and the performance through the same synthesizer and soundfont, and report the L2 distance between their log-mel spectrograms (Mel-L2) and the Pearson correlation of their frame-wise RMS loudness curves in dB (Loud.\ corr)---whether the loudness contour rises and falls with the score's.

\subsection{Dynamics Realization and Note Accuracy}
\label{sec:capability}

We compare four compute-matched methods: \textbf{RoboPianist}~\cite{zakka2023robopianist}, the prior-work policy $\pi^{\mathrm{rp}}$ trained on its note-accuracy reward without any dynamics term; \textbf{End-to-End}, a single policy trained on the full composite reward of Eq.~\eqref{eq:composite-reward} (i.e., $\pi^{\mathrm{vel}}$ itself); and the residual applied to either frozen base---\textbf{Residual on RoboPianist} and \textbf{Residual on $\pi^{\mathrm{vel}}$} (\method's two-stage procedure). Table~\ref{tab:main} reports aggregate results over \subsetname.

\begin{figure}[!t]
\centering
\includegraphics[width=\linewidth]{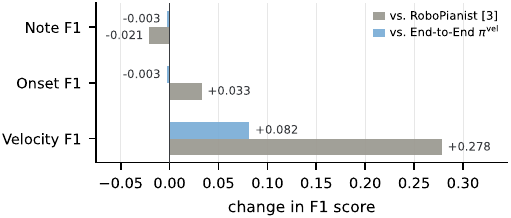}
\caption{\textbf{What the residual stage changes} on \subsetname: change in each F1 metric relative to RoboPianist~\cite{zakka2023robopianist} and to the end-to-end policy trained on the same reward; positive means \method is better. Absolute values are in Table~\ref{tab:main}.}
\label{fig:tradeoff}
\end{figure}

RoboPianist plays most of the notes but few of them at the right loudness---note F1 0.779 against a Velocity F1 of 0.063---the gap that motivates this work.
\method raises Velocity F1 to 0.342, improving every one of the 51 songs, and more than halves velocity MAE (37.1 $\to$ 17.2).

Figure~\ref{fig:tradeoff} separates two conclusions.
First, the residual outperforms training the same objective end-to-end at comparable note accuracy.
Second, even on a base never trained for dynamics it lifts Velocity F1 to 0.198 with almost no loss in note accuracy; the velocity-supervised base is nevertheless required to reach the full improvement.
In audio space the End-to-End policy and the residual on $\pi^{\mathrm{vel}}$ are substantially closer to the reference than RoboPianist, and comparable to each other (Table~\ref{tab:main}).

\subsection{Ablation Studies}
\label{sec:ablation}

\begin{table}[!t]
\centering
\caption{\textbf{Reward-term ablation} (Nocturne, Rousseau transcription; $5{\times}10^6$ steps): each dynamics reward alone, the other coefficient zero, against the pair $(0.2,\,0.5)$.}
\label{tab:reward-ablation}
\small
\setlength{\tabcolsep}{3.5pt}
\begin{tabular}{lcccc}
\toprule
Reward setting & Note F1 $\uparrow$ & Onset F1 $\uparrow$ & Vel MAE $\downarrow$ & Vel F1 $\uparrow$ \\
\midrule
Velocity only & \textbf{0.81} & 0.31 & \textbf{8.8} & 0.19 \\
Onset only    & 0.78 & \textbf{0.49} & 32.8 & 0.06 \\
Both          & 0.79 & 0.44 & 9.5 & \textbf{0.30} \\
\bottomrule
\end{tabular}
\end{table}

\paragraph{Complementary reward components}
The two dynamics rewards address complementary failure modes (Table~\ref{tab:reward-ablation}).
With the velocity reward alone, the policy achieves low velocity error on matched onsets but matches relatively few target events.
Because velocity MAE is evaluated only on matched events, it can remain low even when onset coverage is poor.
Conversely, the onset-coverage reward alone yields the highest onset F1 but provides no supervision for target intensity.
Combining both terms achieves the highest Velocity F1, which requires an event to be correct in both onset and velocity.
This result supports the coverage term as a safeguard against satisfying the velocity objective on only a small subset of target events.

\begin{figure}[!t]
\centering
\includegraphics[width=\linewidth]{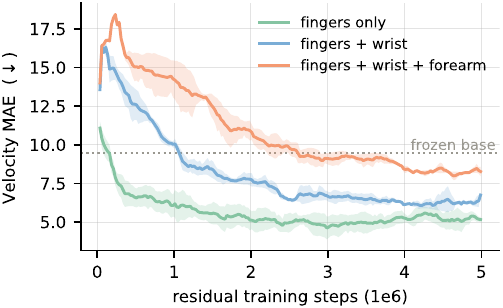}
\caption{\textbf{Residual action space} (Nocturne, Rousseau transcription; $5{\times}10^6$ residual steps, 3 seeds, mean $\pm$1 s.d.; dotted line = the frozen base). Widening the residual beyond the fingers slows learning and converges to a higher velocity error.}
\label{fig:dof}
\end{figure}

\paragraph{Residual action-space localization}
We examine whether the residual should be restricted to the fingers or extended to the wrist and arm (Figure~\ref{fig:dof}).
The finger-only residual rapidly reduces velocity error below that of the frozen base and converges to the lowest final error.
In contrast, expanding the action space slows learning: the wider residuals spend much of training only approaching the frozen-base error, and their final Velocity F1 is lower (0.35 with the wrist and 0.33 with the forearm, against 0.41 for fingers only).
These results support the intended control decomposition, in which the base preserves proximal arm and wrist coordination while the residual adapts distal finger motions for strike intensity.
We therefore use the finger-only residual throughout.


\subsection{Generalization Across Intensity Commands}
\label{sec:controllability}
\begin{figure}[!t]
\centering
\includegraphics[width=\linewidth]{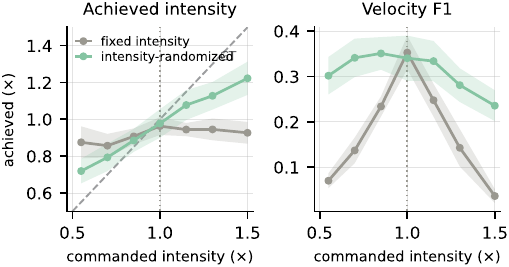}
\caption{\textbf{Commanded intensity control} (4 songs, mean $\pm$ s.e.m.; the reference is the same policy trained at a fixed intensity). Both axes are multiples of the song's written loudness; the dashed diagonal is exact realization. \emph{Left:} achieved intensity. \emph{Right:} Velocity F1 across the command range.}
\label{fig:control-curve}
\end{figure}

We test whether a single policy can follow an intensity command at run time, using the end-to-end policy $\pi^{\mathrm{vel}}$ trained with the intensity randomization described in Section~\ref{sec:mdp}.
The reference uses the same policy and training budget but a fixed training intensity.
We use the end-to-end policy because randomization was less effective with the frozen-base residual, which achieved approximately two thirds of the modulation gain.
At evaluation, we sweep the command from $0.5\times$ to $1.5\times$ on four songs (Figure~\ref{fig:control-curve}).
This range matches the training support; extrapolation beyond it is not evaluated.
Commanded and realized intensities are normalized by each song's written intensity to compare the song-specific policies on common axes.

\begin{figure}[!t]
\centering
\includegraphics[width=\linewidth]{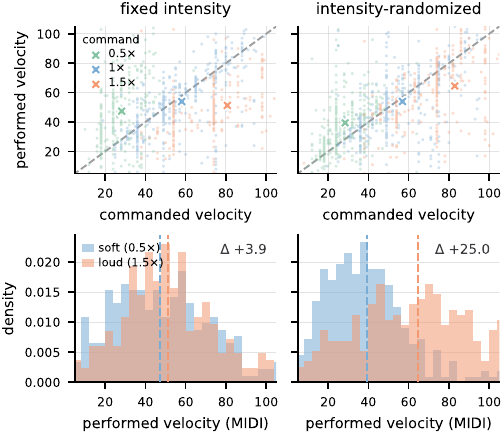}
\caption{\textbf{Per-note behavior under commanded intensity} (matched onsets, 4 songs). \emph{Top:} performed against commanded velocity under three commands (dashed: command realized exactly; $\times$: each command's mean)---the means climb along the diagonal only under randomization. \emph{Bottom:} velocity densities under the softest and loudest command (dashed: means): the command shifts the bulk of the distribution, though the two still overlap.}
\label{fig:control-hist}
\end{figure}

The randomized policy responds monotonically to the command (slope ${+}0.54$), whereas the fixed-intensity reference remains nearly constant.
Velocity F1, computed against the commanded velocities $g_t^{\mathrm{cmd},(k)}$, remains usable across the range, and note F1 is nearly unchanged.
Figure~\ref{fig:control-hist} further shows that the realized-velocity distribution shifts by approximately one and a half notated dynamic levels between the softest and loudest commands, a shift absent from the reference.
The loudest command is only partially realized, however.
These results demonstrate in-range generalization across intensity commands, rather than exact tracking or extrapolation beyond the training range.

\subsection{Robustness to Design and Evaluation Choices}
\label{sec:robustness}
We test whether our conclusions remain stable under changes to training, inference, and evaluation settings.

\begin{figure}[!t]
\centering
\includegraphics[width=\linewidth]{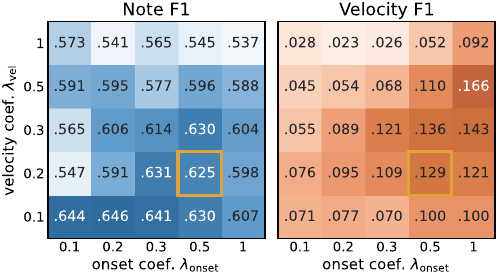}
\vspace{3pt}

\small
\setlength{\tabcolsep}{3.5pt}
\begin{tabular}{lcccc}
\toprule
$(\lambda_{\mathrm{vel}},\lambda_{\mathrm{onset}})$ & Note F1 $\uparrow$ & Onset F1 $\uparrow$ & Vel MAE $\downarrow$ & Vel F1 $\uparrow$ \\
\midrule
$(0.2,\,0.2)$ & 0.78\,\tiny$\pm$0.01 & 0.41\,\tiny$\pm$0.01 & 25.5\,\tiny$\pm$2.0 & 0.14\,\tiny$\pm$0.01 \\
$(0.2,\,0.5)$\,$\star$ & \textbf{0.79\,\tiny$\pm$0.01} & \textbf{0.49\,\tiny$\pm$0.05} & 25.5\,\tiny$\pm$2.7 & \textbf{0.16\,\tiny$\pm$0.03} \\
$(0.3,\,0.5)$ & 0.76\,\tiny$\pm$0.01 & 0.43\,\tiny$\pm$0.01 & 23.6\,\tiny$\pm$2.0 & 0.16\,\tiny$\pm$0.01 \\
$(0.5,\,0.5)$ & 0.73\,\tiny$\pm$0.01 & 0.35\,\tiny$\pm$0.02 & \textbf{21.0\,\tiny$\pm$1.8} & 0.15\,\tiny$\pm$0.02 \\
\bottomrule
\end{tabular}
\caption{\textbf{Reward coefficients} (three songs each, named in Section~\ref{sec:robustness}). \emph{Top:} $5{\times}5$ screening grid at $3{\times}10^6$ steps, one seed per cell; the outlined cell is the setting we use. \emph{Bottom:} the four settings the grid cannot separate, re-run at $5{\times}10^6$ steps with three seeds ($\pm$~s.d. over seeds; $\star$ = adopted).}
\label{fig:pareto}
\end{figure}

\paragraph{Reward weights}
We first screen $(\lambda_{\mathrm{vel}},\lambda_{\mathrm{onset}})$ over a $5{\times}5$ grid at $3{\times}10^6$ steps with a single seed per cell, averaged over three songs---\emph{Nocturne} Op.~9 No.~2 (a transcription distinct from the Rousseau \emph{Nocturne} of Table~\ref{tab:reward-ablation}), \emph{Waltz} Op.~69 No.~2, and \emph{Norwegian Dance} Op.~35 No.~3 (Figure~\ref{fig:pareto}). The two axes pull against each other: note accuracy is highest where the velocity weight is smallest, which is exactly where dynamics is weakest, and the cells with the best Velocity F1 give up note accuracy, while the largest velocity weight is worst on both. We therefore re-run the four settings the grid cannot separate for $5{\times}10^6$ steps on three songs---\emph{Clair de Lune}, \emph{Waltz} Op.~69 No.~2, and \emph{Norwegian Dance} Op.~35 No.~3---with three seeds per setting and song (Figure~\ref{fig:pareto}, bottom). The extremes separate by more than the seed spread: raising $\lambda_{\mathrm{vel}}$ to $0.5$ costs $0.06$ note F1 and $0.14$ onset F1 against s.d.\ of at most $0.01$ and $0.05$, and lowering $\lambda_{\mathrm{onset}}$ to $0.2$ costs onset and Velocity F1 without buying note accuracy. Velocity MAE runs the other way---it is lowest at $\lambda_{\mathrm{vel}}{=}0.5$, the setting with the worst coverage---which is the behavior Section~\ref{sec:setup} warns about: measured on matched onsets only, it improves when difficult notes are dropped, whereas Velocity F1 does not. We adopt $(0.2,0.5)$, which leads on the metrics that count every scored note; a larger $\lambda_{\mathrm{vel}}$ buys little additional dynamics and costs note accuracy, and we prefer a setting that leaves the base task intact.

\begin{figure}[!t]
\centering
\includegraphics[width=\linewidth]{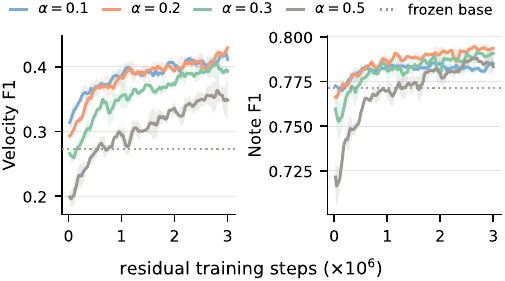}
\caption{\textbf{Residual scale $\alpha$} (\emph{Prelude} Book~1 No.~2, three seeds, mean $\pm$1 s.d.): a separate residual is trained at each $\alpha$ on the frozen $\pi^{\mathrm{vel}}$ base, tracked over residual training. Dotted lines are the frozen base.}
\label{fig:alpha-sweep}
\end{figure}

\paragraph{Residual scale}
The scale $\alpha$ bounds how far the residual may move the base action, so it is fixed before training: we train a separate residual at each $\alpha$ on \emph{Prelude} Book~1 No.~2 with three seeds (Figure~\ref{fig:alpha-sweep}). Every scale improves on the frozen base, but the useful range is $\alpha{\leq}0.2$; beyond it Velocity F1 falls steadily, to $0.341{\pm}0.027$ at $\alpha{=}0.5$. Within that range note accuracy is left undisturbed: note F1 stays above the frozen base, and at $\alpha{=}0.2$ it is also the steadiest across seeds ($0.793{\pm}0.004$). We adopt $\alpha{=}0.2$, the larger of the two scales the sweep cannot separate: it gives the residual as much authority as it can take while note execution stays intact.

\begin{table}[!t]
\centering
\caption{\textbf{Velocity F1 under different tolerances $\tau$} (mean over \subsetname). The ranking is identical at every $\tau$; we report $\tau{=}8$.}
\label{tab:tau}
\small
\setlength{\tabcolsep}{3.5pt}
\begin{tabular}{lcccc}
\toprule
Method & $\tau{=}4$ & $\tau{=}8$ & $\tau{=}12$ & $\tau{=}15$ \\
\midrule
RoboPianist~\cite{zakka2023robopianist} & 0.034 & 0.063 & 0.092 & 0.114 \\
End-to-End ($\pi^{\mathrm{vel}}$)  & 0.160 & 0.260 & 0.319 & 0.348 \\
Residual on RoboPianist            & 0.133 & 0.198 & 0.232 & 0.248 \\
\midrule
Residual on $\pi^{\mathrm{vel}}$   & \textbf{0.251} & \textbf{0.342} & \textbf{0.377} & \textbf{0.392} \\
\bottomrule
\end{tabular}
\end{table}

\paragraph{Evaluation tolerance}
Velocity F1 uses a tolerance of $\tau{=}8$, corresponding to half a notated dynamic level (Section~\ref{sec:setup}). Repeating the main comparison for $\tau{=}4$--$15$ preserves the ranking of all four methods (Table~\ref{tab:tau}). Moreover, the residual policy has its largest relative margin over the End-to-End policy at the strictest tolerance, showing that the comparison does not depend on the chosen $\tau$.

\begin{figure}[!t]
\centering
\includegraphics[width=\linewidth]{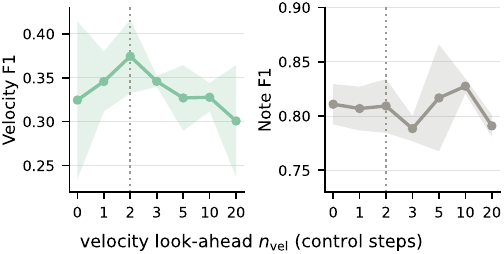}
\caption{\textbf{Velocity look-ahead} (Nocturne, Rousseau transcription; $8{\times}10^6$ steps, 3 seeds, mean $\pm$1 s.d.; dotted line = the setting we use). Note accuracy is on its own scale.}
\label{fig:lookahead}
\end{figure}

\paragraph{Velocity look-ahead}
We vary the look-ahead horizon from the current target only ($n_{\mathrm{vel}}{=}0$) to $n_{\mathrm{vel}}{=}20$ (Figure~\ref{fig:lookahead}). A short horizon is enough: Velocity F1 is highest at $n_{\mathrm{vel}}{=}2$, and seeing further ahead does not help---longer horizons fall back to the level of $n_{\mathrm{vel}}{=}0$ or slightly below---while note accuracy is unaffected throughout. We use $n_{\mathrm{vel}}{=}2$. Overall, \method remains effective across moderate variations in its design and evaluation settings.

\section{Limitations}
\label{sec:limitations}
Velocity is the only expressive dimension we model; tempo and articulation~\cite{borovik2023scoreperformer,jeong2019virtuosonet} remain open.
The dynamics module does not yet generalize across repertoire: a residual distilled from many per-song specialists preserved note accuracy on its training songs, but on unseen songs any velocity gain came at the cost of missed notes.
Finally, all results are in simulation: we have not deployed on a real piano, and Eq.~\eqref{eq:vel-map} is a simulation-calibrated approximation of a nonlinear physical action~\cite{russell1998hammer}. Sim-to-real transfer remains a central challenge in this domain~\cite{zeulner2025realpiano,xie2026handelbot}.

\section{Conclusion}
\label{sec:conclusion}
We presented \method, a dynamics-aware framework for robotic piano performance: a score-to-contact interface that makes keystroke intensity observable, coverage-aware event objectives that make it learnable without sacrificing notes, and a structure-preserving finger-only residual that adds it to a frozen base policy.
On \subsetname\ this yields substantially more faithful dynamics at a small, quantified cost in note accuracy, and a single intensity-conditioned policy follows a commanded intensity at run time.
Beyond piano, the same approach---exposing a sensing channel, defining an event-level reward, and attaching a residual---may extend expressive objectives to other pretrained dexterous controllers.


\bibliographystyle{IEEEtran}
\bibliography{references}

\end{document}